\def\paperID{XXXX}
\def\confName{XXXX}
\def\confYear{XXXX}

\def\authorBlock{
Antonio Alliegro$^{1,2}$\qquad
Yawar Siddiqui$^{3}$\qquad
Tatiana Tommasi$^{1}$\qquad
Matthias Nie{\ss}ner$^{3}$\\
Politecnico di Torino$^1$ \qquad
Italian Institute of Technology$^2$\\
Technical University of Munich$^3$
}

\newif\ifreview 
\newif\ifarxiv \newcommand{\arxiv}{\arxivtrue}
\newif\ifcamera 
\newif\ifrebuttal 

\arxiv %
\pdfoutput=1
\documentclass[10pt,twocolumn,letterpaper]{article}
\ifreview \usepackage[review]{cvpr} \fi
\ifarxiv \usepackage[pagenumbers]{cvpr} \fi
\ifrebuttal \usepackage[rebuttal]{cvpr} \fi
\ifcamera \usepackage{cvpr} \fi

\usepackage{graphicx}	
\usepackage{amsmath}	
\usepackage{amssymb}	
\usepackage{booktabs}
\usepackage{times}
\usepackage{microtype}
\usepackage{epsfig}
\usepackage[table,xcdraw,dvipsnames]{xcolor}
\usepackage{caption}
\usepackage{float}
\usepackage{placeins}
\usepackage{color, colortbl}
\usepackage{stfloats}
\usepackage{enumitem}
\usepackage{tabularx}
\usepackage{xstring}
\usepackage{multirow}
\usepackage{xspace}
\usepackage{url}
\usepackage{subcaption}
\usepackage{xcolor}
\usepackage[hang,flushmargin]{footmisc}

\ifcamera \usepackage[accsupp]{axessibility} \fi

\newcommand{\mypara}[1]{\vspace{3pt}\noindent\textbf{#1}}

\ifarxiv  \fi

\newcommand{\R}[1]{{%
    \textbf{%
        \ifstrequal{#1}{1}{\textcolor{red}{R#1}}{%
        \ifstrequal{#1}{2}{\textcolor{blue}{R#1}}{%
        \ifstrequal{#1}{3}{\textcolor{magenta}{R#1}}{%
        \ifstrequal{#1}{4}{\textcolor{teal}{R#1}}{%
                           \textcolor{cyan}{R#1}%
        }}}}%
    }%
}}

\usepackage{xr-hyper}

\makeatletter
\newcommand*{\addFileDependency}[1]{
  \typeout{(#1)}
  \@addtofilelist{#1}
  \IfFileExists{#1}{}{\typeout{No file #1.}}
}

\makeatother

\definecolor{cvprblue}{rgb}{0.21,0.49,0.74}
\usepackage[pagebackref,breaklinks,colorlinks,citecolor=cvprblue]{hyperref}
\usepackage[capitalize]{cleveref}
\crefname{section}{Sec.}{Secs.}
\crefname{table}{Table}{Tables}
\crefname{figure}{Fig.}{Figs.}

\usepackage{graphicx}
\usepackage{amsmath}
\usepackage{amssymb}
\usepackage{booktabs}
\usepackage{multirow}
\usepackage{comment}
\usepackage{xspace}
\usepackage{xcolor}
\usepackage{bm}
\usepackage{amsbsy}
\usepackage{pgfplots}
\pgfplotsset{compat=1.17}
\definecolor{darkblue}{rgb}{0.28, 0.24, 0.55}

\usepackage[pagebackref,breaklinks,colorlinks]{hyperref}

\newcommand{\OURS}{PolyDiff\xspace}
\definecolor{soft_blue}{RGB}{135,206,250}
\definecolor{light_gray}{RGB}{240,240,240}

\begin{document}
\title{\OURS: Generating 3D Polygonal Meshes with Diffusion Models}
\author{\authorBlock}
\twocolumn[{
    \renewcommand\twocolumn[1][]{#1}%
    \maketitle
    \begin{center}		
    \vspace{-5mm}
    \includegraphics[width=\linewidth]{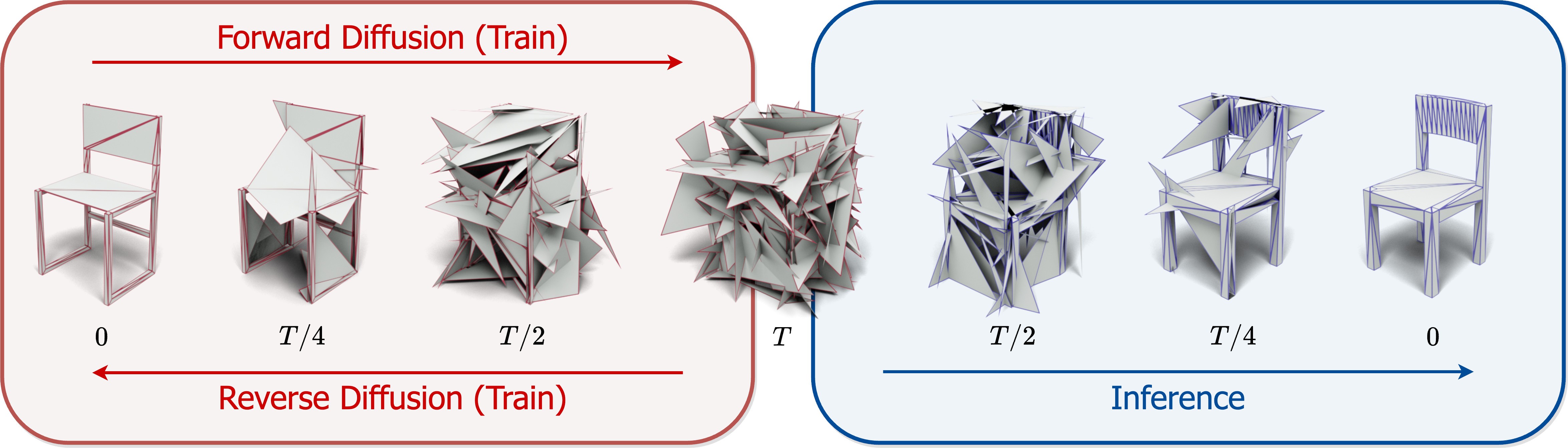}
    \captionof{figure}{We propose \OURS, a novel 3D generative approach that operates natively on polygonal meshes. 
    Meshes are treated as quantized triangle soups which are progressively corrupted with categorical noise. The process is then reverted by a transformer-based denoiser trained to restore the original mesh. \OURS is the first diffusion-based model able to generate realistic and diverse 3D polygonal meshes.}
    \label{fig:teaser}
    \end{center}    
}]

\maketitle
\begin{abstract}
We introduce \OURS, the first diffusion-based approach capable of directly generating realistic and diverse 3D polygonal meshes. 
In contrast to methods that use alternate 3D shape representations (\eg implicit representations), our approach is a discrete denoising diffusion probabilistic model that operates natively on the polygonal mesh data structure. 
This enables learning of both the geometric properties of vertices and the topological characteristics of faces. 
Specifically, we treat meshes as quantized triangle soups, progressively corrupted with categorical noise in the forward diffusion phase. 
In the reverse diffusion phase, a transformer-based denoising network is trained to revert the noising process, restoring the original mesh structure. 
At inference, new meshes can be generated by applying this denoising network iteratively, starting with a completely noisy triangle soup. 
Consequently, our model is capable of producing high-quality 3D polygonal meshes, ready for integration into downstream 3D workflows. 
Our extensive experimental analysis shows that \OURS achieves a significant advantage (avg. FID and JSD improvement of 18.2 and 5.8 respectively) over current state-of-the-art methods.
\vfill
\end{abstract}

\section{Introduction}
\label{sec:intro}
Creating 3D content is a challenging task with a wide range of applications in movies, video games, design modeling, robotics, and virtual reality. 
Currently, generating high-fidelity 3D shapes and scenes, requires extensive effort from skilled 3D artists who manually craft them. 
This labor-intensive process can largely benefit from the assistance of learning-based generative models, which can streamline and enhance the creation workflow. 

In the context of 3D shape modeling, the focus is primarily on surfaces composed of polygons, favored for their suitability in visualization, editing, and texturing. Modern graphics pipelines are optimized for mesh processing, but meshes present unique complexities for generative machine learning models as they are non-Euclidean structured and irregular. 
They consist of vertices freely positioned in 3D space and faces of varying sizes, with the number of both vertices and faces being variable. This irregularity poses additional challenges for deep learning methods, which are typically designed to operate on regular, grid-based data structures as images. 

Existing works navigate these challenges by adopting alternate representations more suitable to learning methods, such as voxels~\cite{3dGANvoxel2016,zhou21_pvd}, point clouds~\cite{groueix2018,luo21_pointcloud_diff,zeng22_lion}, and distance fields~\cite{Shim_2023_CVPR,impl2018Chen}. However, adapting data to fit these formats often leads to a loss of the unique characteristics intrinsic to the original mesh representation, \eg lack of sharp edges and planar surfaces. Additionally, the outputs from these methods usually require conversion back into meshes for downstream 3D content consumption. This conversion, typically done through postprocessing steps like marching cubes~\cite{marchingcubes} or other isosurface extraction methods, often results in over-tessellated or over-smoothed outputs, thus unable to achieve the quality of compact and high-fidelity meshes created by experts.

To this end, we propose \OURS (see Figure~\ref{fig:teaser}), a method that directly generates 3D shapes as polygonal meshes using Denoising Diffusion Probabilistic Models~\cite{sohl2015_deep, ho2020_denoising}. In our approach, meshes are represented as triangle soups with discrete coordinates. During the noising phase, these coordinates are incrementally corrupted with categorical noise. We train a denoising transformer to reverse this noise on the categorical labels at any diffusion timestep. For inference, novel meshes can be generated by starting with fully noised triangle soups and applying the denoising process iteratively over a fixed number of steps. This discrete diffusion process aligns well with the inherently discrete characteristics of polygonal meshes. \OURS produces meshes that are not only cleaner but also more varied than those generated by existing state-of-the-art methods, showing an average FID gain of 18.2 and JSD gain of 5.8 with respect to the best-performing baseline.

\smallskip
\noindent To sum up, our contributions are as follows:
\begin{itemize}[leftmargin=*]
\item We introduce a novel formulation for mesh generation, adapting discrete denoising diffusion models to operate on mesh data. 
This is achieved by representing meshes as a quantized triangle soup, a comprehensive data structure encapsulating both the vertices and the faces of the 3D mesh. 
\item Our method is evaluated on the task of unconditional mesh generation, where it outperforms current state-of-the-art approaches by producing significantly cleaner and more coherent results. 
\end{itemize}

\section{Related Work}
\label{sec:related}
Generative models have witnessed significant improvements in the past few years, especially thanks to the development of Denoising Diffusion Probabilistic Models (DDPMs)~\cite{sohl2015_deep, ho2020_denoising}. Rather than learning to generate samples in a single forward pass as in the case of variational autoencoders (VAE, \cite{Kingma2014_VAE}) and generative adversarial networks (GAN, \cite{NIPS2014_GAN}), diffusion models exploit iterative denoising. A gradual corruption turns the data distribution into noise and then a model is trained to revert this process with multiple steps, each of which reconstructs a small amount of the information that the corruption removed. The success of these models has been particularly evident in boosting image generation quality \cite{dhariwal2021diffusion, Rombach_2022_CVPR} and supporting techniques in visual computing \cite{po2023state}.

The largest part of the literature on DDPMs focuses on Gaussian diffusion processes (the noise is an isotropic Gaussian distribution) that operate in continuous state spaces. This prevents their application on discrete data as text and structural data as graphs with categorical nodes and edge attributes. 
One of the initial formulations employing discrete state spaces for text applications was introduced in \cite{austin2021_structured}: this approach involved novel corruption processes to shape data generation and incorporated auxiliary losses to stabilize the training process. 
An equivariant denoising diffusion model that operates on atom coordinates and categorical features was also defined in \cite{hoogeboom2022_equivariant} to generate molecules in the 3D space. 
More recently, DiGress \cite{vignac2023_digress} proposed to progressively edit graphs with a specific type of noise that consists of adding or removing edges and changing their categorical attributes. 
In \cite{Shabani23_housediffuse}, structured geometry data as floorplans are generated through a diffusion model that combines continuous and discrete denoising: it reorders the position of a set of polygonal elements that come with specific inter-relations. A guided noise injection was also proposed in 
\cite{chen23_polydiffuse} to better control the positioning of each element in the set of polygons. The focus for those works is on 2D data and their application to 3D is not straightforward. 

Early work on 3D shape generation leverages autoencoder architectures, with the seminal work AtlasNet \cite{groueix2018} proposing to deform primitive surface elements into a 3D shape and mainly building on point clouds.  
DDPMs have been also extended to 3D shape generation starting from point clouds, where the points are considered as particles that are diffused to a noise distribution under heat. Specifically, the authors of \cite{luo21_pointcloud_diff} modeled the reverse diffusion process as a Markov chain conditioned on shape latent that follows a prior distribution parameterized via normalizing flows, and the training objective was derived from the variational bound of the likelihood of point clouds. 
Another pioneering work proposed to combine a denoising diffusion model with the point-voxel 3D shape representation for unconditional shape generation and multiple 3D completion from a partial observation \cite{zhou21_pvd}.
The method in \cite{zeng22_lion} improved over the previous ones by operating on point clouds that are first mapped into regularized latent spaces and then fed to a diffusion model. The latent representation is hierarchical and combines global shape with point-structured information. 

In literature, meshes are often regarded as a secondary output derived through post-processing techniques, such as the application of marching cubes~\cite{marchingcubes} on implicit functions. 
Discretizing the 3D space is an alternative strategy to generate meshes. Specifically, the authors of \cite{liu23_MeshDiffusion} proposed to train a diffusion model on a tetrahedral grid structure where each vertex has a Sign Distance Field (SDF) value. The grid is deformable and thus capable of capturing finer geometric details than what could be done with marching cubes as already discussed in \cite{gao20_deftet}. Still, the learning process passes through multiview RGBD images rather than directly building on raw mesh data structures and requires ad-hoc regularization terms to obtain reasonable results. 

Only a few methods proposed to generate directly 3D meshes directly but none of them leverage DDPMs. 
BSP-Net \cite{chen20_bsp} learns instead an implicit field from point coordinates and a shape feature vector. It exploits plane equations corresponding to binary partitions of space, and their final grouping results in shapes with a markedly cuboidal and fragmented appearance. 
PolyGen \cite{nash20_polygen} employs an autoregressive approach for 3D mesh generation. The process involves two consecutive stages: the first stage unconditionally models mesh vertices, while the second stage models mesh faces based on the input vertices. 
However, the two-stage generation process of PolyGen might restrict the model's flexibility as it does not seamlessly align with the inherent characteristics of 3D meshes, particularly the intricate interplay between mesh vertices and their topological arrangement into faces. 
In contrast, our \OURS represents 3D meshes using a quantized triangle soup data structure and seamlessly models the joint distribution of mesh vertices and faces in a single stage.

\section{Background}
\label{sec:background}
\noindent\textbf{Diffusion Models.}  DDPMs are based on two main components: a noise model, and a denoising neural network. The noise model $q$ is progressively applied to the sample $\mathbf{x}_0$, corrupting it into a sequence of increasingly noisy latent variables $\mathbf{x}_{1:T}=\mathbf{x}_1,\mathbf{x}_2,\ldots,\mathbf{x}_t$. The process is  Markovian so it holds $q(\mathbf{x}_{1:T}|\mathbf{x}_0)=\prod_{t=1}^Tq(\mathbf{x}_t|\mathbf{x}_{t-1})$. The denoising network parametrized by $\theta$ will learn to reverse the process so that $p_\theta(\mathbf{x}_{0:T})=p(\mathbf{x}_T)\prod_{t=1}^Tp(\mathbf{x}_{t-1}|\mathbf{x}_t)$. 

The loss used to evaluate how well the output of the generative model $p_\theta(\mathbf{x}_0)$ fits the data distribution $q(\mathbf{x}_0)$ is based on a variational upper bound for the negative log-likelihood of the data $\mathbf{x}$:
\begin{align}
\label{equation: vlb}
    & \mathcal{L}_{\text{vlb}} = \mathbb{E}_{q} [D_\text{KL}(q(\mathbf{x}_T \vert \mathbf{x}_0) \parallel p_\theta(\mathbf{x}_T))] \nonumber \\
    & + \mathbb{E}_{q} [\sum_{t=2}^T D_\text{KL}(q(\mathbf{x}_{t-1} \vert \mathbf{x}_t, \mathbf{x}_0) \parallel p_\theta(\mathbf{x}_{t-1} \vert\mathbf{x}_t, t))] \nonumber \\
    & - \log p_\theta(\mathbf{x}_0 \vert \mathbf{x}_1),
\end{align}
where $\mathbb{E}_{q}(\cdot)$ denotes the expectation over the joint distribution $q(\mathbf{x}_{0:T})$.

However, this model can be difficult to train as reconstructing the whole chain of variables to turn back some noise $\mathbf{x}_T$ into a realistic sample requires applying the network recursively. The training becomes efficient only under specific conditions which are all satisfied when the noise is Gaussian.  
More specifically, if the transition probability in the forward process is formalized as a Gaussian distribution 
\begin{equation}
    q(\mathbf{x}_t\vert\mathbf{x}_{t-1})=\mathcal{N}(\mathbf{x}_t;\sqrt{1-\beta_t}\mathbf{x}_{t-1},\beta_t\mathbf{I}),
\end{equation}
with $\{\beta_t\in(0,1)\}_{t=1}^T$ defining the noise schedule, 
and similarly it holds for the reverse process 
\begin{equation}
    p_\theta(\mathbf{x}_{t-1}\vert \mathbf{x}_t)=\mathcal{N}(\mathbf{x}_{t-1};\mu_\theta(\mathbf{x}_t,t),\Sigma_\theta(\mathbf{x}_t,t)),
\end{equation}
then the KL divergence terms in the loss are tractable and can be computed analytically.
Furthermore, via a simple re-parametrization the learning objective reduces to minimizing the L2 loss between the predicted noise vector and the ground truth applied noise. 

\noindent\textbf{Discrete Diffusion Models.} 
In the discrete data domain, each element $x_t^{d=1,\ldots,D}$ of the vector $\mathbf{x}_t$ represents a discrete random variable that can assume one value among $C$ distinct categories~\cite{austin2021_structured, hoogeboom2022_equivariant, vignac2023_digress}. Here, the forward transition probability from state $i$ to state $j$ can be represented by matrices $[\mathbf{Q}_t]_{i,j}=q(x_t=j\vert x_{t-1}=i)$. Moreover, by denoting the one-hot version of $x^d_t$ with the row vector $\boldsymbol{x}^d_t$, we can write
\begin{equation}
    q(\boldsymbol{x}_t^d\vert \boldsymbol{x}_{t-1}^d)=\texttt{Cat}(\boldsymbol{x}_{t}^d;\boldsymbol{p}=\boldsymbol{x}_{t-1}^d\mathbf{Q}_t),
\end{equation}
where $\texttt{Cat}(\boldsymbol{x}^d_t;\mathbf{p})$ is a categorical distribution over the one-hot vector $\boldsymbol{x}^d_t$, with probabilities given by the row vector $\mathbf{p}$. Here, $\boldsymbol{x}_{t-1}^d\mathbf{Q}_t$ should be understood as a row vector-matrix product, and the transition matrix $\mathbf{Q}_t$ is applied on each element $\boldsymbol{x}_t^d$ independently, with $q$ that factorizes on these higher dimensions as well. 

In the following, we will drop the superscript $d$ for simplicity, and write $q(\boldsymbol{x}_t\vert \boldsymbol{x}_{t-1})$ to refer to a single $d$-th element of $\mathbf{x}_t$ and $\mathbf{x}_{t-1}$. Starting from $\boldsymbol{x}_0$, the following $t$-step marginal is 
\begin{equation}
    q(\boldsymbol{x}_t\vert \boldsymbol{x}_{0})=\texttt{Cat}(\boldsymbol{x}_{t};\mathbf{p}=\boldsymbol{x}_{0}\overline{\mathbf{Q}}_t),
\end{equation}
with $\overline{\mathbf{Q}}_t=\mathbf{Q}_1\mathbf{Q}_2\cdots\mathbf{Q}_t$, and the posterior at time $t-1$ is 
\begin{align}
\label{equation: cat_q}
    q(\boldsymbol{x}_{t-1}\vert & \boldsymbol{x}_t,\boldsymbol{x}_0)= \frac{q(\boldsymbol{x}_t \vert \boldsymbol{x}_{t-1}, \boldsymbol{x}_0) q(\boldsymbol{x}_{t-1} \vert \boldsymbol{x}_0) }{ q(\boldsymbol{x}_t \vert \boldsymbol{x}_0)} \nonumber\\ 
    =\texttt{Cat}&\left (\boldsymbol{x}_{t-1};\mathbf{p}=\frac{\boldsymbol{x}_t\mathbf{Q}_t^\top\odot\boldsymbol{x}_0\overline{\mathbf{Q}}_{t-1}}{\boldsymbol{x}_0\overline{\mathbf{Q}}_t\boldsymbol{x}_t^\top} \right),
\end{align}
where $\odot$ is element-wise multiplication. 
Considering the Markov property of the forward noise process, it holds $q(\boldsymbol{x}_t\vert\boldsymbol{x}_{t-1},\boldsymbol{x}_0)=q(\boldsymbol{x}_t\vert\boldsymbol{x}_{t-1})$. By factorizing the reverse denoising process so that it is conditionally independent over the $d=1,\ldots,D$ elements, the KL divergence between $q$ and $p_\theta$ can be computed by summing over all possible values of each random variable. 
With a uniform transition matrix $\mathbf{Q}_t$, the corresponding $\overline{\mathbf{Q}}_t$ is easy to compute and the diffusion model training runs efficiently. 
It is also possible to substitute the variational upper bound in Eq. (\ref{equation: vlb}) with a cross-entropy loss function which imposes a slightly stronger form of supervision, comparing the model's predicted probability of each category against the clean input. 

\section{PolyDiff}
\label{sec:method}
\mypara{Data Representation.} 
We tailor DDPMs for mesh generation and operate on its irregular data representation. 
Each mesh consists of a set of 3D vertices and a set of $k$-sided faces. 
More formally, we indicate the set of vertices of a mesh as $V = \{v_i | i = 1, \ldots, n\}$ and the set of faces as $F = \{f_j | j = 1, \ldots, m\}$.
Each face comprises $k$ indices to the set $V$ and defines a single polygon. In this work, we focus on triangle meshes ($k=3$) but %
our approach can seamlessly extend to the domain of quadrangular meshes. 

For a given triangle mesh, each vertex $v \in \mathbb{R}^3$ is represented by a triplet of Euclidean coordinates 
$v_i = (x_i, y_i, z_i)$, while each face $f \in \mathbb{Z}^3$ by a triplet of vertex-indices $f_j = (i_1, i_2, i_3)$. 
By quantizing the space with $2^N$ bins for each coordinate, the vertex set changes into $\tilde{V} = \{ \tilde{v}_{i} = (Z(x_i), Z(y_i), Z(z_i)) \mid i = 1, \ldots, n \}$ where $Z: \mathbb{R} \rightarrow \{0, \ldots, 2^N - 1\}$. Thus, a mesh is represented by a \emph{quantized triangle soup} $T \in \mathbb{Z}^{m \times 3 \times 3}$. 

Vertices and faces jointly determine the 3D object's geometry and surface characteristics, with their interplay being vital to the mesh's structure. Modeling them independently could result in geometric inconsistencies. 
Our quantized triangle soup representation offers a simple yet comprehensive way to encode the complete mesh geometry. 

\begin{figure}[t]
    \centering
    \includegraphics[width=1\linewidth]{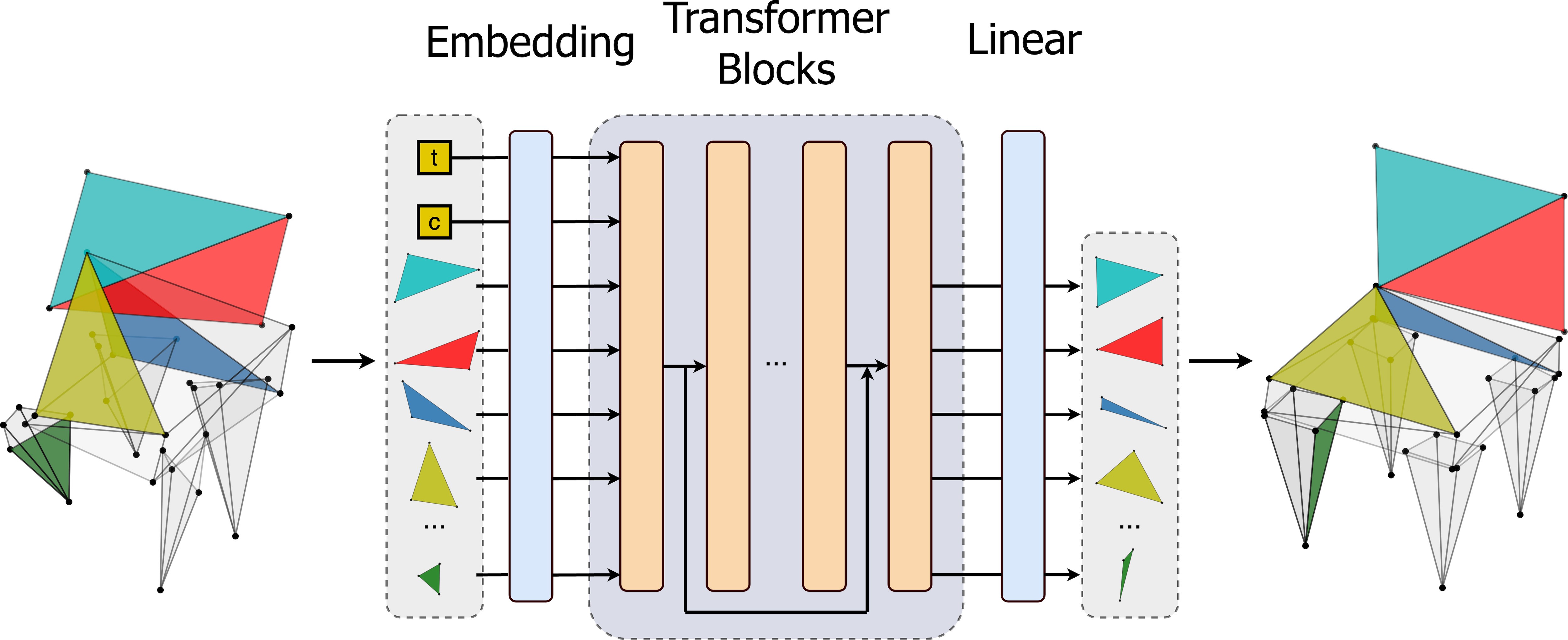}
    \caption{\OURS represents meshes as quantized triangle soups. Each mesh face is composed of three vertices, with each vertex being represented by a triplet of discrete coordinate values. 
    For each mesh, the noising process incrementally alters the categorical values of the vertex coordinates over several timesteps.  
    The noised version of the mesh is input to a transformer network which is tasked with predicting the clean, uncorrupted mesh at any given timestep. 
    The training is driven by a cross-entropy loss between the network's predicted vertex coordinates and those of the original, uncorrupted mesh.
    }
    \label{fig:method_diagram}
    \vspace{-4mm}
\end{figure}

\mypara{Noising Process.} 
Starting from a set of $S$ meshes, we employ a discrete denoising diffusion model to learn the categorical distribution of the triangle soup data $T_s$ with $s=1,\ldots, S$. It operates through a step-by-step modification of the discrete vertex coordinate values. 
In forward diffusion, $T_s$ is subjected to a progressive introduction of noise across a sequence of timesteps. Noise perturbation leads to modifications in the categorical values corresponding to the quantized coordinates of the vertex triplets. 

\mypara{Denoising Process.} In reverse diffusion, a deep neural network is trained to predict the correct categorical value of the coordinates starting from the intermediate noised representation $T_{s}^{t}$ at the timestep $t$. 
At the core of the denoising network (Figure~\ref{fig:method_diagram}) we employ a Vision Transformer (ViT~\cite{dosovitskiy2020_vit}), originally designed for diffusion-based image synthesis applications~\cite{bao2022_uvit}. 
At the timestep $t$ of the reverse diffusion process, the denoising network takes as input the noisy mesh representation $T_{s}^{t}$ and outputs the estimated denoised version $T_{s}^{t-1}$. 
The process initiates with the transformation of the categorical values in $T_{s}^{t}$ into continuous high-dimensional features, yielding an embedded tensor $\mathcal{T}_{s,emb}^{t}$ with dimensions $\mathbb{R}^{m \times 3 \times 3 \times D}$. 
After undergoing positional encoding and linear projection, these high-dimensional features are aggregated per face, 
resulting in a condensed matrix $\mathcal{F}_s^t$ in $\mathbb{R}^{m \times C}$. This matrix is finally fed to the transformer alongside the diffusion timestep $t$ and the object-category label to predict the configuration of the denoised mesh $T_{s}^{t-1}$.

\mypara{Loss Function.} 
As a loss function, we adopt the cross entropy computed between the predicted denoised mesh version and the original uncorrupted $T_s$. 

We denote the cross-entropy loss at each 3D vertex of the mesh as $\mathcal{L}^v = -\sum_{h=1}^{3}\sum_{c=0}^{2^N-1} p_{h,c}\log(\hat{p}_{h,c})$,
where $h$ ranges over the three coordinates of the vertex, $p$ represents the ground truth one-hot distribution of quantized vertex coordinates extracted from $T_s$, and $\hat{p}$ denotes the predicted probabilities of denoised coordinates across the $2^N$ quantized bins. 
Consequently, the overall loss is expressed as:
\begin{equation}
\mathcal{L}_{\text{ce}} = \sum_{j=1}^m\sum_{k=1}^3{\mathcal{L}^v_{j,k}}~,\nonumber
\end{equation}
As a result, we formalize the task of mesh generation as a classification problem that can be scaled up with modern learning frameworks.

\mypara{Implementation Details.} 
Our model is implemented in PyTorch and trained for a maximum of 2000 epochs on 8 NVIDIA A100 GPUs, with a batch size of 128 for each GPU. The training process spans 4 days. 
We employ AdamW~\cite{loshchilov2018_adamw} optimizer with a learning rate of 5e-4. The learning rate is modulated using a cosine annealing schedule throughout the training. Additionally, we incorporate a warmup phase during the initial 200 epochs. 
In our experiments, we use 1000 diffusion timesteps and adopt a cosine noise schedule~\cite{nichol2021_improved}. 
During training, all meshes are normalized such that the length of the long diagonal of the mesh bounding box is equal to 1. After normalization, the vertices undergo an 8-bit quantization process. 
We train our model on multiple augmented versions of each 3D training model, obtained by varying the planar decimation angle. 
Additionally, we apply random scaling by independently sampling scaling factors for each axis within the range of $[0.75, 1.25]$. 
In the training phase, we learn a unique embedding for each object category, which is fed into the diffusion model as a conditioning. This approach allows us to produce examples that match a specific category during the inference stage. 

\begin{table*}[h]
\centering
\resizebox{0.9\textwidth}{!}{
\begin{tabular}{lcccccc}
\toprule
\rowcolor{light_gray}
\textbf{Category} & \textbf{Model} & \textbf{MMD ($\downarrow$)} & \textbf{COV (\%, $\uparrow$)} & \textbf{1-NNA (\%, $\downarrow$)} & \textbf{JSD ($\downarrow$)} & ~~~~~~~\textbf{FID ($\downarrow$)} \\
\midrule
\multirow{5}{*}{Chair}
        & AtlasNet~\cite{groueix2018}  & 18.08 & 22.42 & 83.03 & 54.87 & ~~~~~~~209.48 \\
        & BSPNet~\cite{chen20_bsp}   & \textbf{16.26} & 33.45 & 75.21 & 22.81 & ~~~~~~~73.86 \\
        & PolyGen~\cite{nash20_polygen}& 23.74	& 37.09	& 86.61	& 71.26	& ~~~~~~~48.27 \\
        & \OURS     & 18.57 & \textbf{49.58} & \textbf{58.67} & \textbf{14.69} & ~~~~~~~\textbf{41.07} \\ 
        \cmidrule{2-7}
        & Train     & 16.01 & 53.58 & 50.36 & 11.60 & ~~~~~~~11.00 \\
\hline

\multirow{5}{*}{Table}
        & AtlasNet~\cite{groueix2018} & 18.16 & 20.36 & 86.28 & 34.31 & ~~~~~~~186.05 \\
        & BSPNet~\cite{chen20_bsp}  & 15.33 & 33.43 & 76.35 & 20.97 & ~~~~~~~63.57 \\
        & PolyGen~\cite{nash20_polygen}& 19.19 & 40.92 & 74.10 & 56.56 & ~~~~~~~46.15 \\
        & \OURS     & \textbf{15.16} & \textbf{50.60} & \textbf{57.14} & \textbf{13.12} & ~~~~~~~\textbf{26.17} \\
        \cmidrule{2-7}
        & Train     & 13.59 & 56.69 & 47.26 & 10.90 & ~~~~~~~10.79 \\
\hline

\multirow{5}{*}{Bench}
        & AtlasNet~\cite{groueix2018} & 11.05 & 35.06 & 72.99 & 123.58 & ~~~~~~~222.90 \\
        & BSPNet~\cite{chen20_bsp}  & \textbf{10.33} & 36.21 & 73.28 & 101.99 & ~~~~~~~85.69 \\
        & PolyGen~\cite{nash20_polygen}  & 16.21 & 41.95 & 80.46 & 143.60 & ~~~~~~~81.90 \\
        & \OURS     & 11.44 & \textbf{43.68} & \textbf{61.49} & \textbf{86.60} & ~~~~~~~\textbf{49.72} \\
        \cmidrule{2-7}
        & Train     & 9.83 & 51.72 & 49.14 & 70.11   & ~~~~~~~44.80 \\
\hline

\multirow{5}{*}{Display}
        & AtlasNet~\cite{groueix2018} & 14.12 & 43.54 & 67.01 & 74.49 & ~~~~~~~197.60 \\
        & BSPNet~\cite{chen20_bsp}  & \textbf{13.35} & 38.78 & 68.71 & \textbf{61.09} & ~~~~~~~76.82 \\
        & PolyGen~\cite{nash20_polygen} & 17.61 & 44.90 & 62.93 & 81.59 & ~~~~~~~56.03 \\
        & \OURS     & 15.28 & \textbf{45.58} & \textbf{56.80} & 69.02 & ~~~~~~~\textbf{42.60} \\  
        \cmidrule{2-7}
        & Train     & 13.36 & 60.54 & 45.92 & 59.84 & ~~~~~~~28.15 \\
\bottomrule
\end{tabular}
}
\caption{Quantitative evaluation of the performance of PolyDiff with respect to PolyGen, BSPNet, and AtlasNet. We recall that only PolyGen directly works on mesh vertices and faces as \OURS. 
`Train' refers to the metrics computed using the training samples rather than the generated data. 
Our \OURS outperforms prior state-of-the-art both on the point cloud metrics and perceptual quality.}
\label{tab:quantcomparison}
\end{table*}
\begin{table*}[h]
\centering
\resizebox{0.8\textwidth}{!}{%
\begin{tabular}{lcccccc} 
\toprule
\rowcolor{light_gray}
\textbf{Chair} & \textbf{Diffusion steps} & \textbf{MMD ($\downarrow$)} & \textbf{COV (\%, $\uparrow$)} & \textbf{1-NNA (\%, $\downarrow$)} & \textbf{JSD ($\downarrow$)} \\
\midrule

\multirow{3}{*}{\OURS~Small}
        & 300  & 19.36 & 51.39 & 64.00 & 16.98 \\
        & 500  & 19.83 & 47.88 & 64.91 & 17.83 \\
        & 1000 & 19.55 & 49.70 & 64.42 & 14.65 \\
        & 2000 & 20.69 & 50.91 & 68.48 & 19.85 \\
        \hline

\multirow{3}{*}{\OURS}
        & 300  & 19.04 & 52.24 & 60.97 & 15.11 \\
        & 500  & \textbf{18.67} & \textbf{52.48} & \textbf{59.82} & 14.47 \\
        & 1000 & 18.96 & 50.67 & 60.30 & \textbf{14.33} \\
        & 2000 & 19.97 & 49.94 & 64.06 & 16.69 \\
        \hline

\multirow{1}{*}{Continuous Diff.}
        & 1000  & 23.33 & 34.55 & 75.58 & 85.49 \\
        \hline

\multirow{1}{*}{Train}
        & - & 16.01 & 53.58 & 50.36 & 11.60 \\   
\bottomrule
\end{tabular}
}
\caption{Analysis of the impact of the number of diffusion steps and denoising network model size for the Chair category. 
Additionally, we conduct a comparative analysis between our discrete diffusion approach (\OURS) and continuous diffusion applied to 3D triangle soups.}
\label{tab:ablation}
\vspace{-3mm}
\end{table*}

\begin{figure}[t]
    \centering
    \includegraphics[width=0.9\linewidth]{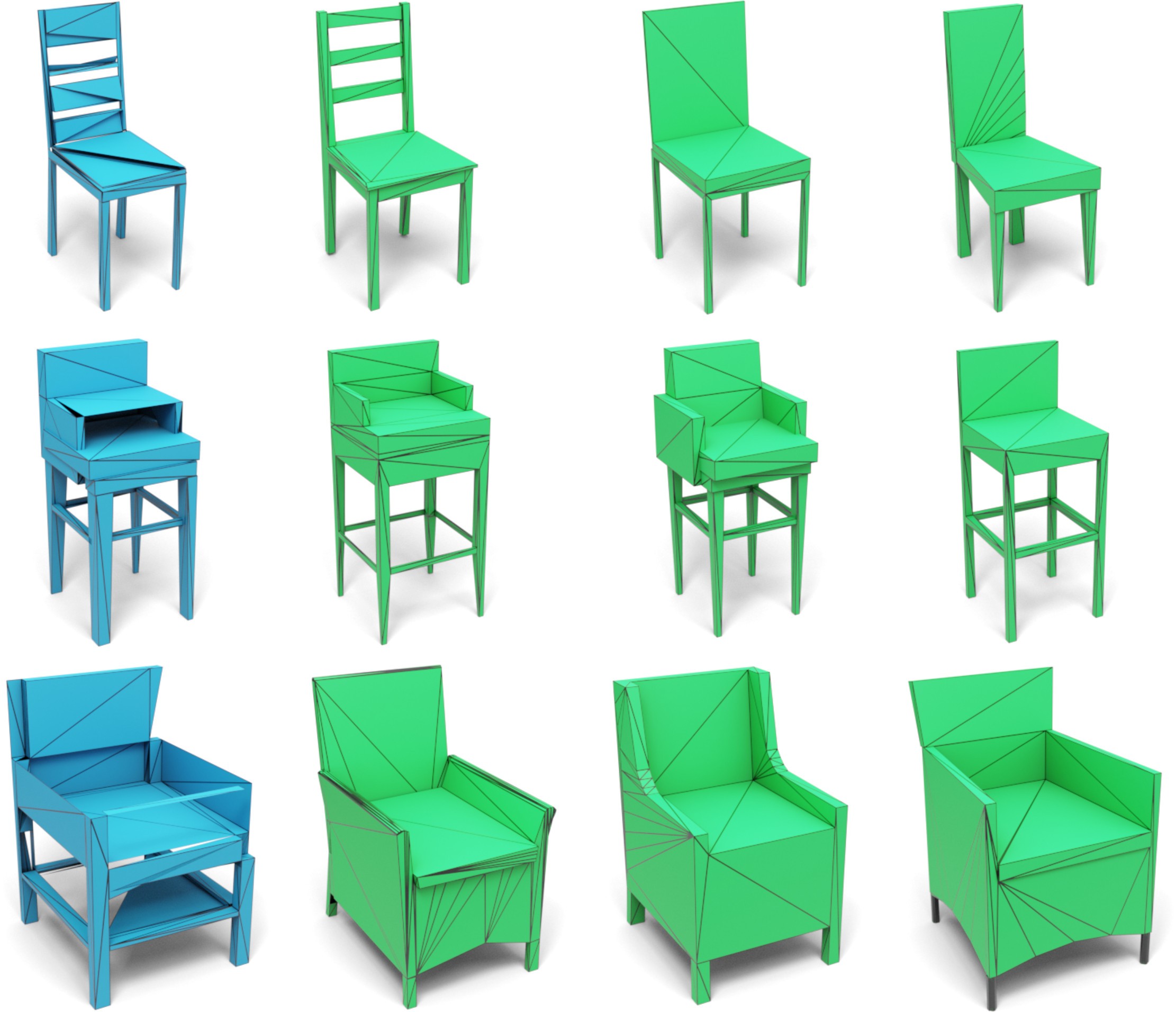}
    \caption{Novel shape generation vs nearest neighbor retrieval. Each row shows one sample generated by our \OURS (left, blue mesh) and the top-3 nearest neighbors (green) from the training set based on the Chamfer distance. The examples show that our method can generate novel shapes.}
    \label{fig:novelty_anal}
    \vspace{-4mm}
\end{figure}

\begin{figure*}[htb!]
    \centering
    \includegraphics[width=0.95\linewidth]{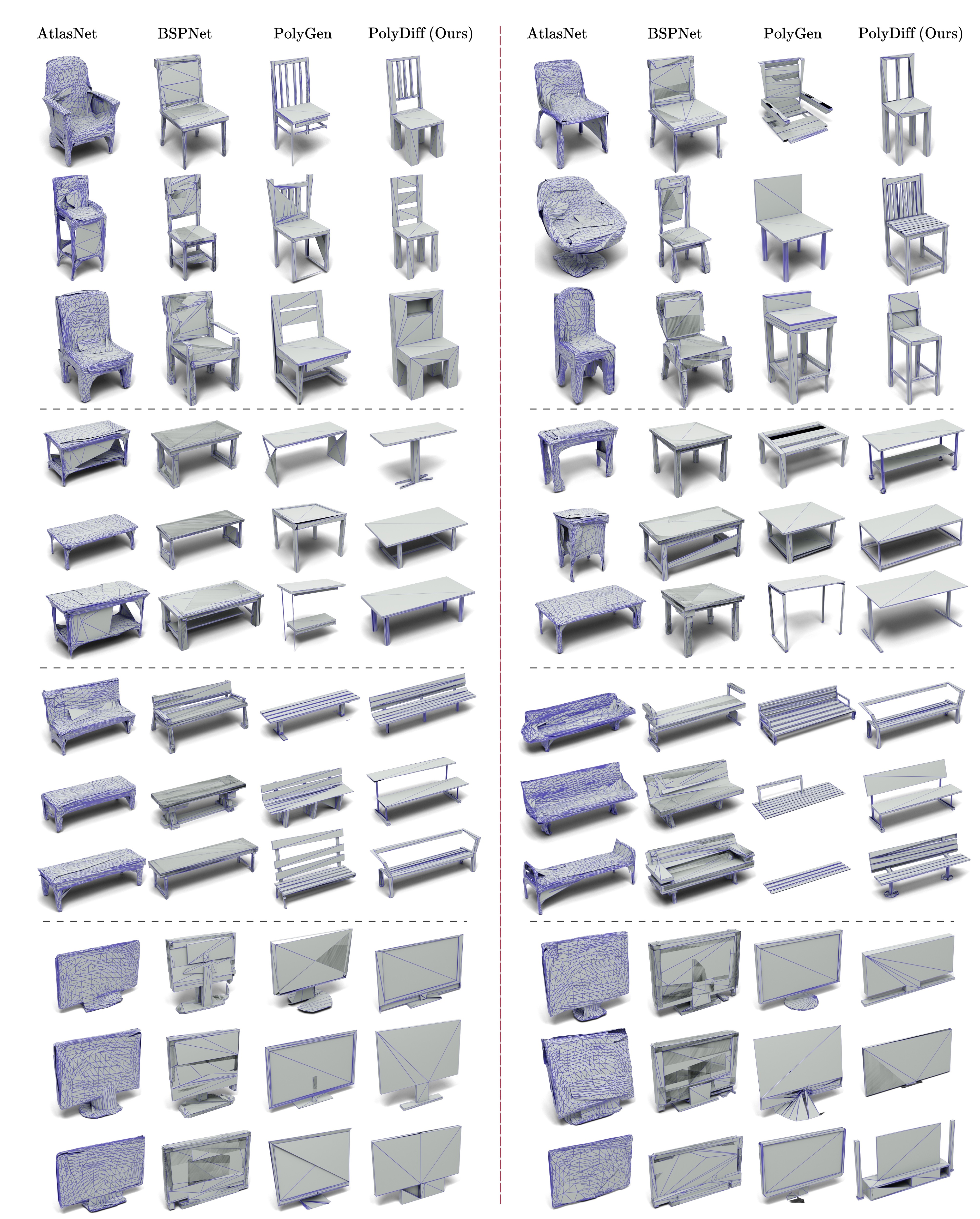}
    \caption{Qualitative comparison of generated meshes. The instances generated by \OURS are noticeably cleaner and more realistic than those of the competitors that show small or overlapping triangles, artifacts, and missing parts.}
    \label{fig:qual_comparison}
\end{figure*}

\begin{figure*}[h]
    \centering
    \includegraphics[width=0.9\linewidth]{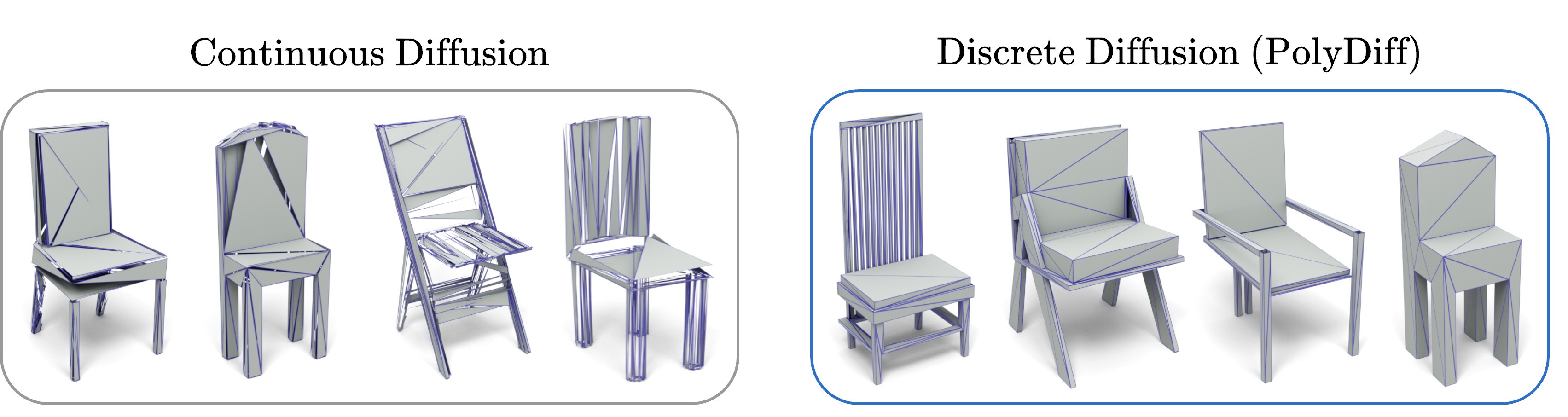}
    \caption{Qualitative comparison between our \OURS discrete diffusion model and a variant that employs standard continuous Gaussian noise. The continuous version exhibits noticeably poorer performance, affirming our hypothesis that discrete diffusion is better suited to the inherently discrete characteristics of mesh data.}
    \label{fig:continuous_diff}
    \vspace{-3mm}
\end{figure*}

\section{Experiments}
\label{sec:experiments}
We evaluate \OURS{} for unconditional generation of 3D meshes. 
In the following, we detail our experimental setup before presenting and discussing the results of our method.

\subsection{Experimental Setup}
\noindent\textbf{Dataset.}
We use the ShapeNet~\cite{chang2015_shapenet} dataset, focusing on four categories: chairs, benches, displays, and tables. The meshes are pre-processed via planar decimation which reduces the number of vertices and faces while combining thin, elongated triangles. 
All the instances are elaborated in this way at 30 different angles, ranging from 1 to 60 degrees, to produce multiple decimated versions of the original mesh. 
We then calculate the Hausdorff Distance for each produced version compared to the original one to identify and discard cases where the decimation process had negatively impacted the model's integrity. 
Finally, we assembled a diverse collection of 3D models, comprising 2746 chairs, 576 benches, 487 displays, and 3340 tables.  
Each unique model in this collection comes with a variety of augmented versions given by the variation of the decimation angle. 
We allocated 90\% of the models to the training set and the remaining 10\% to the testing set, maintaining this ratio across all object categories. 
The models that did not meet our criteria post-decimation are set aside for validation purposes. The maximum number of faces is set to 800.

\noindent\textbf{Evaluation Metrics.} 
Assessing the quality of unconditional generation is challenging due to the absence of a one-to-one mapping between generated samples and ground truth data. 
Following the methodology of previous studies~\cite{luo21_pointcloud_diff,zhou21_pvd,zeng22_lion,achlioptas2018_gan}, we assess the quality of our synthesized shapes using four metrics: Minimum Matching Distance (MMD), Coverage (COV), 1-Nearest-Neighbor Accuracy (1-NNA) and the Jensen-Shannon Divergence (JSD). 
The MMD score measures the fidelity of the generated samples and the COV score detects mode-collapse. 
The 1-NNA metric uses a nearest-neighbor classifier to differentiate between generated and reference samples. 
The JSD score measures the divergence between the point distributions of the generated set and the reference set. 
For MMD and JSD, a lower score indicates better performance. Conversely, for COV, a higher score denotes better coverage, and for 1-NNA, a rate of 50\% represents the optimal value. 
MMD and JSD scores are multiplied by a factor of $10^3$. 
All these metrics are computed on 2048-dim point clouds uniformly sampled from meshes in the generated set and a reference test set. Point clouds are normalized into a bounding box of $[-1, 1]^3$, so that the metrics focus on the shape of the 3D object rather than scale. As a point-set distance metric we adopt the Chamfer distance~\cite{fan2017_chamfer}. 
Additionally, we consider the Frechet Inception Distances (FID)~\cite{heusel2017_gans}. 
FID is calculated using views rendered from the meshes and serves as an indicator of the quality of the generated surfaces. 
For each mesh, we render eight views from distinct camera positions uniformly spaced across the surface of a sphere. 
The FID score is computed using features extracted from each view with an Inception-v3 model pre-trained on the ImageNet~\cite{imagenet} dataset.

\subsection{Results}
\noindent\textbf{Comparison to the state of the art.} We evaluate the unconditional generation abilities of \OURS comparing it with AtlasNet~\cite{groueix2018}, BSPNet~\cite{chen20_bsp}, and PolyGen~\cite{nash20_polygen}. 
For AtlasNet and BSPNet, both of which are based on autoencoder (AE) architectures, we adopt the methodology described in \cite{achlioptas2018_gan} and fit a family of Gaussian Mixture Models (GMMs) on the latent space learned by the AE. 
We then sample latent codes from this GMM-fitted distribution and pass them to the decoder component of the AE to generate a mesh. 
We generate 1,000 examples for each category across all methods and then compute the generative metrics with the test set of our dataset as a reference. 
We also consider calculating these metrics by comparing the training samples, instead of the generated data, with the reference test set. The results of this approach are presented and labeled as `Train' in the tables. 
The quantitative results are presented in Table \ref{tab:quantcomparison} and show that \OURS achieves improved performance over the competitors in all shape categories and all metrics except MMD, according to which \OURS behaves slightly worse than BSPNet, but similarly to AtlasNet and always better than PolyGen which is the only competitor directly working on mesh vertices and faces like \OURS. 
We remark that the reliability of MMD has been questioned in the past as it lacks sensitivity to low-quality results \cite{zeng22_lion}. Table \ref{tab:quantcomparison} also presents the FID metric which captures the perceptual quality of our synthesized shapes and confirms the advantage of \OURS over the other methods. 

As a qualitative evaluation, in Figure~\ref{fig:qual_comparison} we conduct a comparative analysis of the generated meshes, spanning across all categories and methods. 
The meshes generated by AtlasNet are over-triangulated and often exhibit self-intersecting issues. In comparison to our method, AtlasNet also appears to have difficulties in accurately defining thin structures, such as chairs' legs or armrests. 
BSPNet generates meshes by combining convex pieces to form the final mesh. The generated meshes often have a blocky appearance and, at times, exhibit an unnatural triangulation pattern. 
The meshes produced by PolyGen are often incomplete, this happens because of the autoregressive network employed which often produces an End Of Sequence token prematurely. 
In contrast to PolyGen, our method demonstrates superior capability in generating 3D shapes that are both more coherent and cohesive. This advancement stems from our distinct approach of joint learning the vertices and faces of the mesh by utilizing a quantized triangle soup data structure as input to our discrete diffusion model.

\noindent\textbf{Architecture and Design Choices.} 
The transformer architecture is a core component of our model and its capacity might impact the performance. To investigate this aspect we train \OURS on the chair category and compare the results obtained by our default U-ViT-Mid with those of the U-ViT-Small version \cite{bao2022_uvit}. The results in Table \ref{tab:ablation} show that the architecture dimensionality minimally affects the results with the major difference visible only on the 1-NNA metric (52.82 vs 64.42). It is also interesting to note that the number of diffusion steps can be efficiently kept small (500-1000) as increasing it may be detrimental. 

We designed \OURS to adopt discrete diffusion as it perfectly suits the discrete nature of the mesh data structure. 
To validate this hypothesis we also design a model that shares the same denoising network as \OURS but differs in its use of a standard Gaussian diffusion process instead of a discrete diffusion process. 
A comparative analysis of these two models, specifically for the chair category, is presented in Table \ref{tab:ablation}. 
Here, \OURS, operating with 1000 diffusion steps, demonstrates superior performance over its continuous diffusion counterpart. 
This superiority is particularly notable in terms of the JSD metric. Additionally, we include qualitative results in Figure \ref{fig:continuous_diff} where we compare samples generated with Continuous Diffusion objective (left) and Discrete Diffusion objective (right). 
Notably, the meshes produced using the continuous diffusion objective exhibit inconsistent triangles that fail to align properly with one another, leading to a poor-quality output mesh. These experiments further validate our hypothesis of adopting a discrete diffusion process for \OURS. 

\noindent\textbf{Novel Shape Synthesis.} To explore the degree of novelty generation of \OURS we show in Figure~\ref{fig:novelty_anal} our synthesized meshes in comparison to the top three nearest neighbors by Chamfer Distance from the training set. 
The results confirm that our method is not simply memorizing train samples and can generate novel shapes. 

\section{Limitations}
While \OURS{} demonstrates notable advancements in unconditional mesh generation, it does have its limitations.
While we have shown \OURS{} to be effective at generative meshes of single objects, the generation of scene-level meshes remains yet to be explored. 
Additionally, the sampling process in diffusion models is inherently slower compared to feedforward methods, which is an area that could benefit from further optimization. 
Better sampling techniques~\cite{karras2022elucidating} and improved formulations~\cite{song2023consistency} can be adopted to speed up sampling. 

\section{Conclusion}
\label{sec:conclusion}
In this paper, we have introduced a novel probabilistic generative model tailored for polygonal 3D meshes. 
Our method embraces the discrete nature of 3D meshes and employs a discrete diffusion model to learn the joint distribution of mesh vertices and faces. 
Our comprehensive analysis shows that \OURS achieves state-of-the-art performance in the unconditional mesh generation task, demonstrating an average improvement of 18.2 and 5.8 for FID and JSD metrics, respectively, compared to previous methods. 
Our approach can be employed to directly learn from human-crafted mesh data distributions, allowing for the generation of novel and diverse 3D meshes similar to those produced by skilled designers. 
This ability holds the potential to significantly diminish the 3D artists' workload, unleashing creative possibilities that were previously untapped. 

\noindent\textbf{Acknowledgements.} 
This study was carried out within the FAIR - Future Artificial Intelligence Research and received funding from the European Union Next-GenerationEU (PIANO NAZIONALE DI RIPRESA E RESILIENZA (PNRR) – MISSIONE 4 COMPONENTE 2, INVESTIMENTO 1.3 – D.D. 1555 11/10/2022, PE00000013). 
This work was supported by the ERC Starting Grant Scan2CAD (804724). 
Further, we extend our appreciation for the travel support provided by ELISE (GA No. 951847).

{\small
\bibliographystyle{ieeenat_fullname}
\bibliography{references}
}
\end{document}